\documentclass[letterpaper, 10pt, conference]{ieeeconf}  % Comment this line out if you need a4paper

\usepackage{url,lineno,microtype}
\usepackage{amsmath,amsfonts,amssymb}
\usepackage{commath}
\usepackage{graphicx}
\usepackage{cite}

\usepackage{pgfplots}
\usepackage{tikz} 

\pgfplotsset{compat=newest}
\usetikzlibrary{decorations.pathmorphing} % noisy shapes
\usetikzlibrary{fit} % fitting shapes to coordinates
\usetikzlibrary{backgrounds} % drawing the background after the foreground
\usetikzlibrary{external}
\usetikzlibrary{matrix}
\usetikzlibrary{pgfplots.groupplots}
\IEEEoverridecommandlockouts                              % This command is only needed if 
\usepackage{graphics} % for pdf, bitmapped graphics files

\title{\LARGE \bf
Experience Reuse with Probabilistic Movement Primitives
}

\author{Svenja Stark$^{1}$, Jan Peters$^{1,3}$ and Elmar Rueckert$^{1,2}$% <-this % stops a space
\thanks{*This project has received funding from the European Union's Horizon 2020 research and innovation programme under grant agreement No 713010 and No 640554.}% <-this % stops a space
\thanks{$^{1}$Intelligent Autonomous Systems, Technische Universit\"at Darmstadt, Hochschulstr. 10, 64289 Darmstadt, Germany
        {\tt\small svenja@robot-learning.de}}%
\thanks{$^{2}$Institute for Robotics and Cognitive Systems, Universit\"at zu L\"ubeck, Ratzeburger Allee 160, 23538 L\"ubeck, Germany
        {\tt\small rueckert@rob.uni-luebeck.de}
        }%
\thanks{$^{3}$Robot Learning Group, Max-Planck Institute for Intelligent Systems, Max-Planck-Ring 4, 72076 T\"ubingen, Germany {\tt\small mail@jan-peters.net}
	}
}

\begin{document}

\maketitle
\thispagestyle{empty}
\pagestyle{empty}

%%%%%%%%%%%%%%%%%%%%%%%%%%%%%%%%%%%%%%%%%%%%%%%%%%%%%%%%%%%%%%%%%%%%%%%%%%%%%%%%
\begin{abstract}

	Acquiring new robot motor skills is cumbersome, as learning a skill from scratch and without prior knowledge requires the exploration of a large space of motor configurations.
	Accordingly, for learning a new task, time could be saved by restricting the parameter search space by initializing it with the solution of a similar task.
	We present a framework which is able of such knowledge transfer from already learned movement skills to a new learning task. The framework combines probabilistic movement primitives with descriptions of their effects for skill representation. New skills are first initialized with parameters inferred from related movement primitives and thereafter adapted to the new task through relative entropy policy search.
	We compare two different transfer approaches to initialize the search space distribution with data of known skills with a similar effect.
	We show the different benefits of the two knowledge transfer approaches on an object pushing task for a simulated 3-DOF robot. We can show that the quality of the learned skills improves and the required iterations to learn a new task can be reduced by more than 60\% when past experiences are utilized.
\end{abstract}

%%%%%%%%%%%%%%%%%%%%%%%%%%%%%%%%%%%%%%%%%%%%%%%%%%%%%%%%%%%%%%%%%%%%%%%%%%%%%%%%
\section{INTRODUCTION}

Currently, most robots can only execute a fix amount of (pre-defined) movement skills and are thus not able to adapt to their environment by learning new skills on the fly when required. At the same time, when learning a new motor skill with, for example, reinforcement learning, the algorithm usually starts with a wild guess resulting in a large exploration phase leading to unpredictable robot movements and usually resulting in a single learned skill. 
When taking a look at human development instead, we see that at a young age, human babies and infants also perform quite random movements with their bodies in response to changes in the environment (or themselves), but once they have learned basic movement skills, they are able to narrow down the required exploration for solving a new task from randomness to task specific movements~\cite{adolph2017development}. From a robotics point of view, such specific exploration not only enables efficiency in human learning, but also provides some safety as new movements are close to past experienced solutions and therefore neither (over)stretch joints nor lead to collisions with the environment. In machine learning and especially in robot learning, one could also profit from such a transfer of knowledge when learning solutions to new tasks, as in this case as well, it not only reduces the learning time, but also the randomness (and thus possible danger) of explored movements. For scenarios where the exact desired movement can not be demonstrated, which is especially the case for highly precise tasks on a robot, transferring knowledge from other tasks could enable the learning of the desired task.

In the long run, transfer learning can enable robots with a faculty for lifelong learning, such that they can adapt to new tasks or environments. The adaptation will become easier and easier over time thanks to the growing skill knowledge.

\setlength{\belowcaptionskip}{-30pt}

\begin{figure}
	\includegraphics[width=0.5\textwidth]{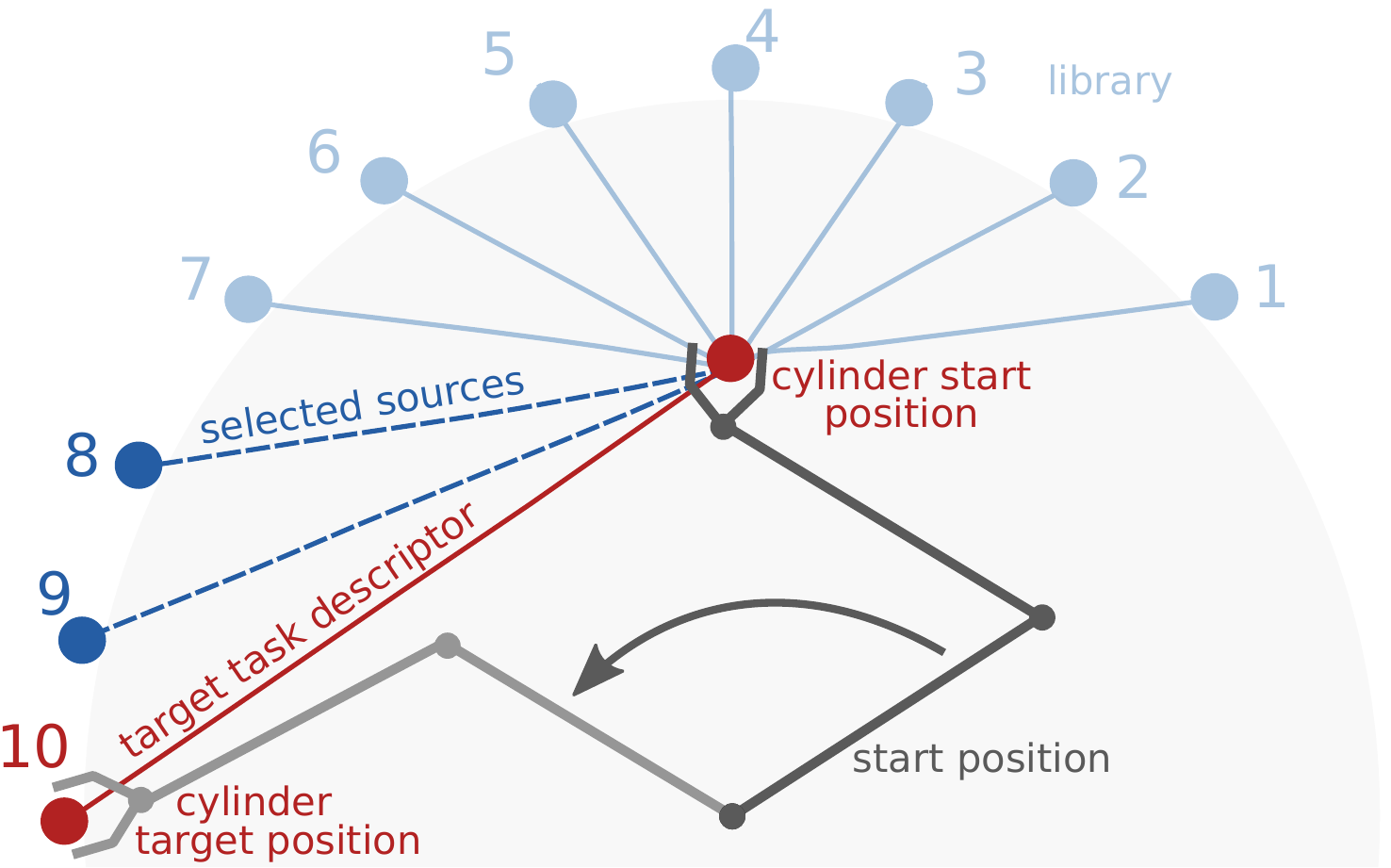}
	\caption{An exemplary depiction of the 2D learning scenario which is used within this work. The robot arm has to solve the new task labeled as $10$ of which it only knows the task descriptor. The task descriptor consists of the desired object trajectory depicted in red. The robot already knows how to solve the other nine depicted skills and has the two closest skills $8$ and $9$ selected to learn from. Their task descriptor is depicted by the dashed line. The workspace of the robot is marked by the gray half-circle. The darker robot arm marks the starting position for training set A and the lighter one marks one possible final position for the given task.}
	%\vspace{-10pt}
	\label{figToyTaskSkills}
\end{figure}

For making some small steps towards closing the gap between human adaptivity and the current state of the art in robotics, we want to enable transfer learning on a robotic setup and therefore present an approach for enabling such transfer on robotic motor skills within this paper. For this purpose,
we combine several existing approaches into a single knowledge transfer framework. We use Probabilistic Movement Primitives (ProMPs)~\cite{NIPS2013_5177}, a probabilistic skill representation which defines a skill as a Gaussian distribution over trajectory parameters. 
Thus, we are able to re-use the mean and the covariance of known source skills to initialize an already pre-shaped and narrowed down search distribution for learning a new task via reinforcement learning. Such an initialization points the learning algorithm towards a promising direction. Furthermore, each skill in our framework consists of a description of its effect alongside the movement trajectory itself. This additional description of the effect, i.e., the task the skill solves, enables the framework to select respective source knowledge for learning a new task and to evaluate the current learning progress.  We assume that such task descriptors for new target tasks are provided by an external planner or a human. Our approach allows us to use knowledge from multiple source skills.

We demonstrate the effectiveness of our framework on a 2D scenario with a 3-DOF planar arm (Figure~\ref{figToyTaskSkills}). We show that using the knowledge of already learned skills to narrow down the parameter search space for learning a new task results in faster convergence and safer exploration.%, which is also safer regarding the explored movements of the robot and leads to more predictable resulting movements. 

\section{Related Work}

The idea of transferring past experience to the process of solving a new task is well known and has already been used extensively \cite{Taylor:2009:TLR:1577069.1755839, Weiss2016, 5288526, HASSABIS2017245} and on a broad variety of learning problems and algorithms such as classification, regression and clustering~\cite{5288526}, deep learning~\cite{DBLP:journals/corr/abs-1808-01974} and reinforcement learning~\cite{Taylor:2009:TLR:1577069.1755839}. Despite its extensive use, there is still a great potential seen in the concept of transfer learning to further propel the machine learning community in different domains~\cite{DBLP:journals/corr/LakeUTG16, youtube} and for robotics, to drive it towards autonomous, lifelong learning robots. % \cite{Oudeyer und Thrun}.

While transfer learning has already been applied within the field of robotics, i.e., for transferring knowledge between robots~\cite{DBLP:journals/corr/HelwaS17, Makondo2018AcceleratingML}, transferring skills from simulation to reality~\cite{DBLP:journals/corr/abs-1809-04720} or policies between robots~\cite{Schwab2018ZeroST}, there has been relatively little research on transferring knowledge from one (motor) skill of a robot to the next, especially in the context of reinforcement learning.
In~\cite{AAMASWS10-barrett}, the value function of SARSA($\lambda$) is transferred when learning a new skill. In contrast, we use a search-distribution-based reinforcement algorithm and transfer the policy search space.  

Such an approach of reusing previously learned policies has been labeled as Policy Reuse~\cite{Fernandez:2006:PPR:1160633.1160762} or, in a broader context, as a starting point method~\cite{torrey2010transfer}. For genetic algorithms, instead of policies, populations have been transferred~\cite{Taylor06transferlearning}. As we use a distribution over polices, our approach does not use concrete offsprings but transfers the whole distribution from which samples in each iteration are drawn and evaluated.

Regarding the usage of task descriptors, there have already been other search distribution methods incorporating features of the desired effect of the skill in the skill learning process, such as e.g. contextual relative entropy policy search~\cite{7781959}.
In contrast to contextual reinforcement learning, our approach is non-parametric and thus does not require a structural dependence between the context and the skill trajectory. Also, as we not only use a final goal position but a full trajectory, we restrict the solution space further, making the resulting trajectories safer and more similar to already known skills.

\section{Skill Library Setup}

In our framework, all acquired skills $S$ of a robot are collected within a skill library $\mathcal{S} = \{S_1, S_2, ..., S_n\}$. Each skill $S_i = (p(\boldsymbol{\tau}_i), \textbf{T}_i)$ is the combination of a movement trajectory distribution $p(\boldsymbol{\tau})$ and a description $\textbf{T}$ of the effect of the movement, i.e., the task or problem it solves. 

Thus, the application of our approach requires a search distribution over the movement trajectories, which, for example, can be realized by parametrized movement representations with a prior distribution over the parameters.
Probabilistic Movement Primitive (ProMP)~\cite{promps_auro} are such kind of representation and we thus use them for each movement trajectory distribution $p(\boldsymbol{\tau})$ as explained further in Section~\ref{sec:promps}.

Each task descriptor $\textbf{T}$ captures the desired or experienced effect of the movement trajectory $\boldsymbol{\tau}$ on the environment, e.g., in the form of the object movement trajectory in task space. Our approach requires that the representation of the task descriptors allows the application of a distance measure:
When receiving the description of a new target task $\textbf{T}^*$ which shall be solved, the most appropriate, i.e. similar, past experience(s) to learn from can be selected by comparing $\textbf{T}^*$ to all known task descriptors $\textbf{T}_{[1..n]}$. 

The movement trajectory distribution(s) of the selected skill(s) can be used to form the initial search distribution for learning the solution $\boldsymbol{\tau}^*$ of the task $\textbf{T}^*$. In this work, we use a policy search algorithm for this purpose. It is further explained in Section~\ref{sec:reps}. The different possibilities of utilizing the knowledge from the skills within the library are presented in Section~\ref{sec:learning}.

\subsection{Movement Trajectory Representation as ProMPs}
\label{sec:promps}

We use ProMPs as probabilistic representations for the movement trajectory, as the probability of this representation directly provides a distribution $p(\boldsymbol{\tau})$. Simple linear approximations would only estimate the weight parameters $\mathbf{w}$ for a linear approximation of a trajectory $\boldsymbol{\tau} = \boldsymbol{\Phi}^\intercal\textbf{w}$ which at most can keep variations over (system) noise $\boldsymbol{\Sigma}_{\boldsymbol{\tau}}$

$$
	p(\boldsymbol{\tau}|\textbf{w}) = \prod_t \mathcal{N} (\tau_t|\boldsymbol{\Phi}_t^\intercal\textbf{w}, \boldsymbol{\Sigma}_{\boldsymbol{\tau}}).
$$

A ProMP additionally captures a variance of the trajectory itself by keeping a Gaussian distribution over the weights $\textbf{w} \sim  \mathcal{N}(\boldsymbol{\mu}_\textbf{w}, \boldsymbol{\Sigma}_\textbf{w})$ and thereby yields a distribution over trajectories. Thus, the probability of observing trajectory $\boldsymbol{\tau}$ is given by

\begin{subequations}
	\begin{align}
		p(\tau_t;\boldsymbol{\theta}) & = \int \mathcal{N} (\tau_t|\boldsymbol{\Phi}_t^\intercal\textbf{w}, \boldsymbol{\Sigma}_{\boldsymbol{\tau}}) \mathcal{N} (\textbf{w}|\boldsymbol{\mu}_\textbf{w}, \boldsymbol{\Sigma}_\textbf{w})d\textbf{w} \nonumber \\
		                              & = \mathcal{N}(\tau_t|\boldsymbol{\Phi}_t^\intercal \boldsymbol{\mu}_\textbf{w}, \boldsymbol{\Phi}_t^\intercal \boldsymbol{\Sigma}_\textbf{w} \boldsymbol{\Phi}_t + \boldsymbol{\Sigma}_{\boldsymbol{\tau}}) \nonumber
	\end{align} %
\end{subequations}
and the open parameters $\boldsymbol{\theta}$ defining a ProMP are $\boldsymbol{\mu}_\textbf{w}$ and $\boldsymbol{\Sigma}_\textbf{w}$, which we estimate when learning a new skill. 

For the features $\boldsymbol{\Phi}$, we use Gaussian basis functions which span over time and which are also linearly distributed over time. Accordingly, each weight $w_i$ mainly influences a certain time span of the trajectory as it weights its respective basis function. As the basis features are the same for all learned trajectories, the weights represent a dimension reduction of the trajectory $\boldsymbol{\tau}$. Such a reduction can be learned for all kinds of trajectories, e.g. trajectories in joint space as well as in task space. Abstract concepts such as being close to an object or orienting the end-effector towards an object are not incorporated within the weights.
It shall be noted that also the task descriptors $\textbf{T}$ can be represented by a ProMP to yield a reduced representation to, for example, compare the learned weights.

To fill a library with $n$ initial skills which can be reused later on, we provide their probabilistic movement representation via learning from demonstration. 
The parameters $\boldsymbol{\mu}_\textbf{w}$ and $\boldsymbol{\Sigma}_\textbf{w}$ can be obtained by providing several demonstrations of the same movement. For each demonstration, we learn the weights $\textbf{w}$ via linear regression or expectation maximization~\cite{Rueckert_ICRA14LMProMPsFinal, srep28455}. Subsequently, we estimate the mean $\boldsymbol{\mu}_\textbf{w}$ and the covariance $\boldsymbol{\Sigma}_\textbf{w}$ via maximum likelihood estimation.

\subsection{Learning a New Skill with REPS}
\label{sec:reps}

To learn a new movement trajectory $\boldsymbol{\tau}^*$ from source knowledge which solves task $\textbf{T}^*$, we use the reinforcement learning algorithm Relative Entropy Policy Search (REPS)~\cite{6439}. More specifically, we employ the episode-based formulation of REPS~\cite{Deisenroth:2013:SPS:2688186.2688187}, which allows a parameterized policy $\pi_\theta(\boldsymbol{\tau})$ to encode a full movement trajectory in an open-loop manner. The reward signal is obtained only after the full execution of one episode.

In our scenario, using a stochastic search algorithm has multiple benefits. First, we can conveniently use the ProMP framework to represent the search distributions of REPS, $\pi_\theta(\boldsymbol{\tau}) = p(\boldsymbol{\tau})$. Hence, the final optimized policy parameters $\theta$ are the parameters of a new ProMP, i.e., a Gaussian distribution. Furthermore, each $\boldsymbol{\tau}$ drawn from policy $\pi_\theta(\boldsymbol{\tau})$ is a linear combination of the learned weights and the basis functions, thus, an executable movement trajectory. This formulation additionally allows to incorporate source knowledge into an initial distribution $q(\boldsymbol{\tau})$, which makes REPS a natural choice for optimizing new movements given prior experience. Second, REPS is a sample-based black-box optimizer, that assumes no knowledge of the accumulated reward function $R(\boldsymbol{\tau})$. This fact is convenient, as even if the desired object trajectory $\textbf{T}^*$ is known, the task descriptor contains no knowledge of the environment, such as (inverse) kinematics or motor and contact dynamics. 

As a policy search algorithm, REPS iteratively optimizes the parameters of the search distribution $\pi_\theta(\boldsymbol{\tau})$ such that the final policy attains maximum expected accumulated reward. The optimization is formulated as a constrained problem for which the Kullback-Leibler divergence (KL) between the optimal policy $\pi_\theta(\boldsymbol{\tau})$ and an initial distribution $q(\boldsymbol{\tau})$ is limited to a threshold $\epsilon$ to limit information loss and greedy updates
\begin{subequations}
	\begin{align}
		\max_{\theta} \quad         & \int \pi_\theta(\boldsymbol{\tau}) R(\boldsymbol{\tau}) \text{d}\boldsymbol{\tau}, \nonumber                               \\
		\text{s.t. } \quad \epsilon & \geq \int \pi_\theta(\boldsymbol{\tau}) \log \frac{\pi_\theta(\boldsymbol{\tau})}{q(\boldsymbol{\tau})} \text{d}\boldsymbol{\tau}, \nonumber \\
		1                           & = \int \pi_\theta(\boldsymbol{\tau}) \text{d}\boldsymbol{\tau}. \nonumber
	\end{align}
\end{subequations}
This optimization is solved by constructing the Lagrangian function and optimizing the dual problem. The resulting optimal new search distribution $\pi_\theta(\boldsymbol{\tau})$ is given by
\begin{equation*}
	\pi_\theta(\boldsymbol{\tau}) \propto q(\boldsymbol{\tau}) \exp(R(\boldsymbol{\tau}) / \eta),
\end{equation*}
which can be seen as a re-weighting of $q(\boldsymbol{\tau})$ based on the performance measure $R(\boldsymbol{\tau})/\eta$, where $\eta$ is the Lagrange multiplier corresponding to the KL constraint.
In practice, this update is performed over the sample set $\{\boldsymbol{\tau}_i, R_i(\boldsymbol{\tau})\}$ as weighted maximum likelihood estimate.

Given that policy reward is only gradually optimized, REPS can implement a safe exploration strategy, by considering local policy updates around a stable trajectory distribution, thus limiting greedy jumps into unfavorable regions in the parameter space. In theory, any other regularized optimization algorithm operating on a (Gaussian) parameter distribution may be used in our framework, such as natural evolution strategies~\cite{wierstra2014natural}, the covariance matrix adaptation evolution strategy~\cite{hansen1996adapting} or random search~\cite{mania2018}.

\begin{figure*}
	\begin{minipage}[t]{\textwidth}
		\includegraphics{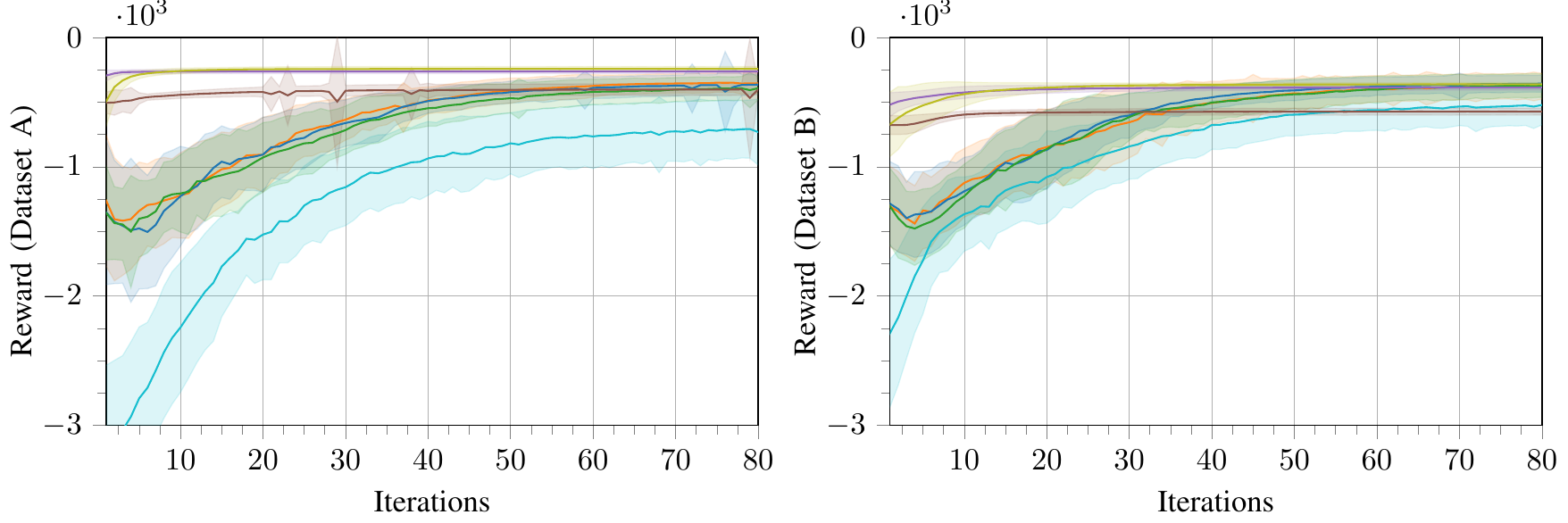}
	\end{minipage}
	\begin{minipage}[t]{\textwidth}
		\centering
		\vspace{-4mm}
		\includegraphics{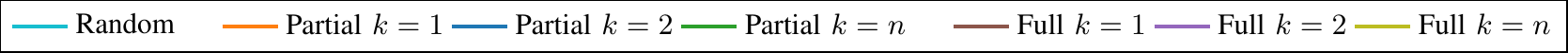}
	\end{minipage}
	\caption{Mean reward per iteration for learning on a large library, averaged over the learning of all ten different skills. The source knowledge is selected from a library containing all skills except the target skill, i.e., $N=9$ for all settings. Such a large library allows k-NN to choose the optimal initialization and it is the optimal setting we introduced within this paper. The label $k=x$ denotes from how many source skills the initialization was compiled. Each iteration evaluates 60 samples to update the covariance matrix and each transfer setting has been repeated $5$ times for all of the ten target skills. The left plot shows the results for learning on dataset A, the right plot shows results for learning on dataset B. The plots include the standard deviation to indicate the variety of the results. The three different initialization modes show different learning characteristics: the random initialization starts with the lowest reward and has the highest variation regarding learned results. The full transfer modes converge fastest, and for $k>1$, they reach the same final reward as the partial transfer. The partial reward outperforms the baseline regarding final reward and higher start for all $k$.}
	%\vspace{-10pt}
	\label{allMaxLibrary}
\end{figure*}

\subsection{Selection and Transfer of Relevant Source Knowledge}
\label{sec:learning}
When getting the description of a new task $\textbf{T}^*$, there are several possibilities to select the appropriate initial policy for REPS to start learning from. As we assume a task descriptor representation $\textbf{T}$ which allows calculation of similarity, we can compare $\textbf{T}^*$ to all familiar task descriptors $\textbf{T}_{[1..n]}$ in the skill library and use the k-nearest neighbors algorithm~(k-NN)~\cite{doi:10.1080/00031305.1992.10475879} to select the $k$ closest skills to learn from in case the library contains more than $k$ skills. For a setting of $k=1$, only the skill with the task descriptor most similar to the target task descriptor is selected to learn from, while for a setting of $k=n$ the whole library knowledge is~utilized.

We propose two different approaches for transferring the selected knowledge. For both approaches, the covariance matrix is scaled using a hand-tuned factor $s$. Otherwise, the search distribution is too restricting to allow the learning of a different movement.  

The first approach is to transfer only the movement trajectory distribution mean of the $k$ selected source skills $\boldsymbol{\mu}_{k}$ and combine it with uniform variance $s\textbf{I}$ as initialization of the REPS search distribution $\mathcal{I}_p = (\boldsymbol{\mu}_{k}, s\textbf{I})$. We call this approach \textit{partial transfer}, and the transferred knowledge could also be gained from other movement representations such as e.g., Dynamical Movement Primitives~\cite{Ijspeert:2013:DMP:2432779.2432781}.

The second approach is to transfer the mean $\boldsymbol{\mu}_{k}$ as well as the scaled covariance $s/(\max(\boldsymbol{\Sigma}_k))\boldsymbol{\Sigma}_k$ of the source skills, leading to a pre-shaped and thus smaller initial search space for REPS $\mathcal{I}_f = (\boldsymbol{\mu}_{k}, s/(\max(\boldsymbol{\Sigma}_k))\boldsymbol{\Sigma}_k)$. We call this second approach \textit{full transfer} and it is only possible with a probabilistic skill representation providing the covariance of the source skill(s).

As a baseline, we also use a `random' parameter initialization which is not based on any source knowledge. To ensure that at least at the beginning of the learning process the robot arm starts in the same position as the other transfer approaches, the weights of the random initialization are a weighted mean between the average over the means of all available skill data $\boldsymbol{\mu}_N = \sum_{n=1}^N \boldsymbol{\mu}_n/N$ and randomly drawn parameters $\boldsymbol{\mu}_r$, thus $\boldsymbol{\mu}_b = \boldsymbol{\lambda} \boldsymbol{\mu}_N + (1-\boldsymbol{\lambda}) \boldsymbol{\mu}_r$. The weights of the average parameters $\boldsymbol{\lambda}$ refer to how large the value of the related basis function $\boldsymbol{\psi}^i$ is at start time $0$, i.e., $\lambda^i = \psi_0^i/\max(\boldsymbol{\psi}_0)$. The baseline is thus set to~$\mathcal{I}_b = (\boldsymbol{\mu}_b, s\textbf{I})$.

\section{Experiments and Evaluation}

We evaluate the presented transfer approach on a scenario of ten object pushing tasks. The task descriptors $\textbf{T}$ are the desired trajectories that the pushed object has to cover and the movement trajectories $\boldsymbol{\tau}$ are joint space trajectories of the~robot. 

The task of pushing an object is non-trivial for an open loop setup, due to the difficulty in modeling the interaction between the robot and the object. A few millimeter difference in contacting the object can result in completely different object trajectories (e.g., the end-effector pushing the object vs. rotating and passing by it). Thus, the robot must not only push along the correct line (which would be a similar task to the end-effector tracking a desired trajectory), but also has to touch the object in the right angle, at the right point, over the whole trajectory. Furthermore, the object has low friction to allow sliding if the arm pushes too forceful.
This interplay also makes the accumulated reward function $R(\boldsymbol{\tau})$ non-linear and thus hard to model, even with the presence of the known target task descriptor, as the interplay between the end-effector and the object is unknown.

One of our goals is safe learning in robotics, and for us, this means that when the robot learns a new movement $\boldsymbol{\tau}^*$, the new movement should have a similar shape to what is known already. Such behavior is more predictable and also sticks to areas of the workspace which are known to be safe. Thus, in addition to a quantitative evaluation regarding the final reward, we evaluate our transfer approaches regarding this safety criterion within a qualitative comparison.

\subsection{Task Setup}

The task uses a 3-DOF planar arm with a gripper and a moveable cylinder in a 2D environment. The dynamics of the environment are simulated by the Bullet engine (PyBullet~\cite{coumans2018}). The arm is controlled in joint space and besides a gripper is mounted to make the pushing easier, the gripper is never closed to actually grasp the object.
%The new movement is learned as a trajectory which is encoded with six basis functions per joint, resulting in overall $18$ parameters
We use $6$ basis functions for each joint trajectory, resulting in overall $18$ parameters for the movement trajectory $\boldsymbol{\tau}^*$. % which have to be learned for a new skill. 
The information available from the environment include the end-effector positions, the cylinder positions and the actual joint positions of the robot during execution. 
The tasks $\textbf{T}^*$ are described by the x and y position of the cylinder at each time step. A snapshot of a possible task learning setup with a large library of nine skills of which two are selected as source knowledge is depicted in Figure~\ref{figToyTaskSkills}. %Data to initilize and were generated by handcrafted demonstrations.
\\

\subsubsection{Datasets}
We evaluate on two different skill datasets, of which both are handcrafted. Dataset A starts with the robot being in a position where the gripper already encompasses the object. In dataset B, the robot starts in a position with a larger distance between the end-effector and the object. A larger distance makes it less likely to randomly push the object at all, but such pushing is necessary for the learning algorithm being able to exploit the respective reward function~component. \\

\begin{figure*}
	\begin{minipage}[t]{\textwidth}
		\includegraphics{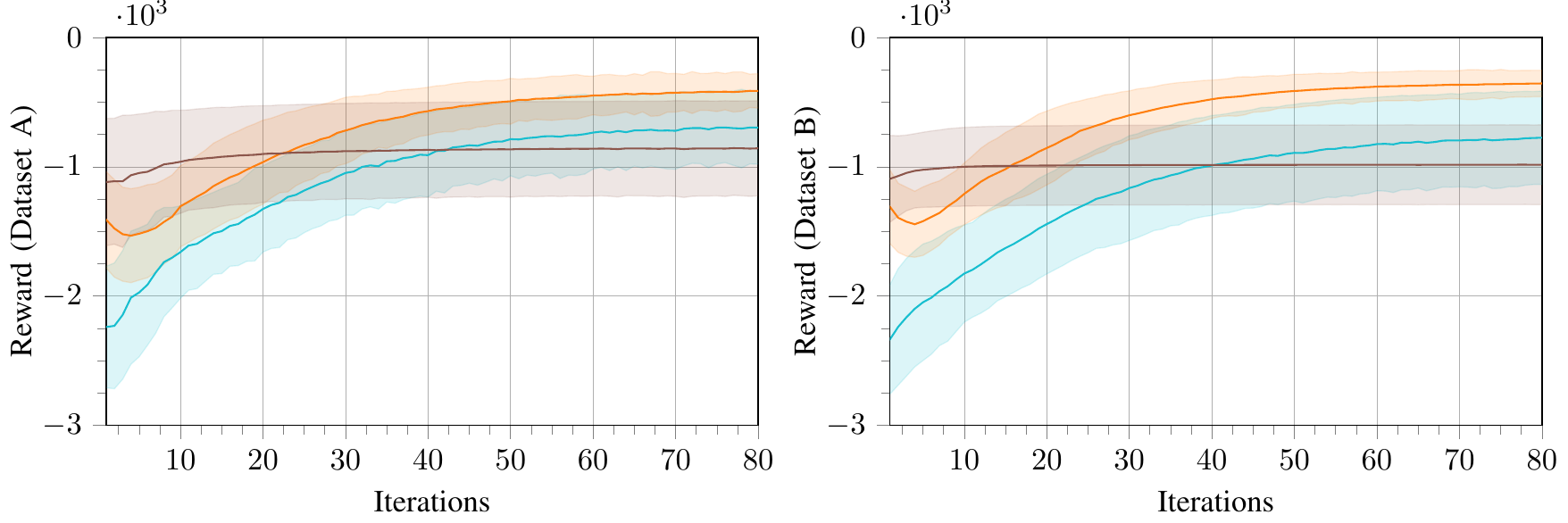}
	\end{minipage}
	\begin{minipage}[t]{\textwidth}
		\vspace{-5mm}
		\includegraphics{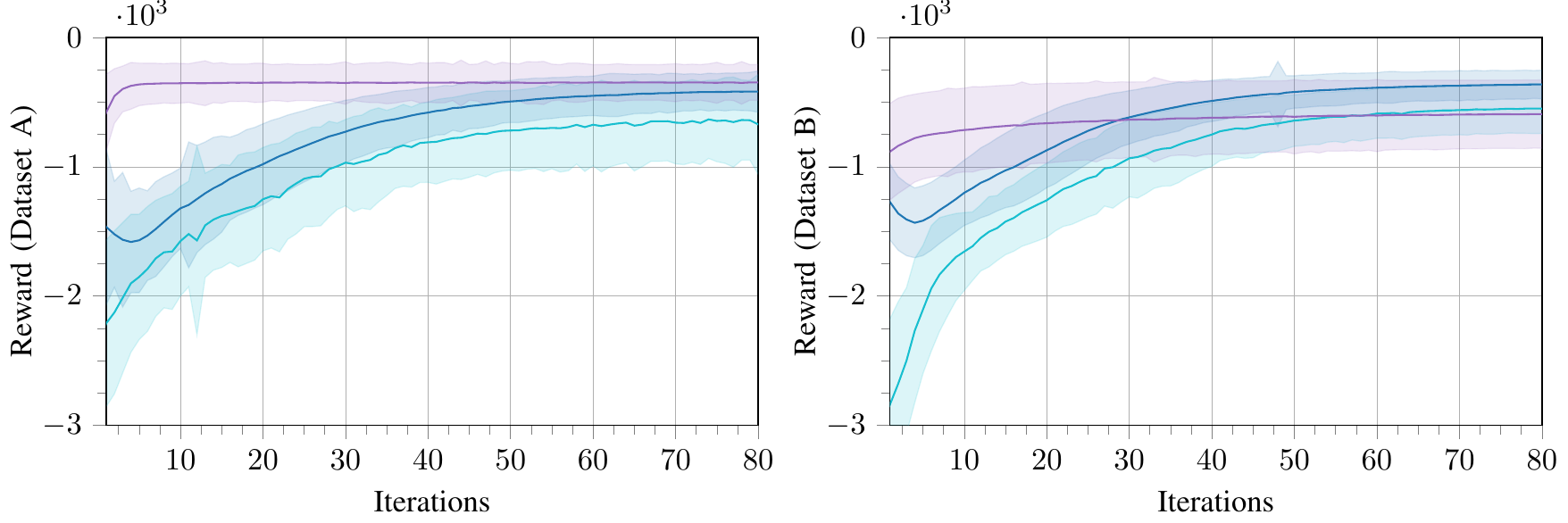}
	\end{minipage}
	\begin{minipage}[t]{\textwidth}
		\centering
		\vspace{-4mm}
		\includegraphics{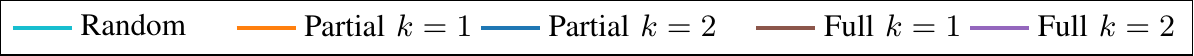}
	\end{minipage}
	\caption{Mean reward per iteration for learning on a small library, averaged over the learning of all ten different skills. The source knowledge is selected from a library containing only the number of skills needed to select $k$ skills, i.e., $N=k$ for all settings. Such a small library annihilates k-NN, as there is no choice possible and thus enforces choosing non-optimal source knowledge. The upper row shows results for learning from a library with size $N=1$, the lower one for learning from a library with size $N=2$. The label $k=x$ denotes from how many source skills the initialization was compiled. The left plot shows the results for learning on dataset A, the right plot shows results for learning on dataset B. The plots include the standard deviation to indicate the variety of the results. The plots show that the partial transfer is more robust to suboptimal source knowledge, as it still yields the same performance as in the previous plot, while the full transfer only manages to outperform the baseline for $k=2$ on dataset A.}
	\label{allMinLibrary}
	\vspace{-1pt}
\end{figure*}

\subsubsection{Task Similarities \& k-NN Modalities}
As we represent our task descriptors $\textbf{T}$ as object trajectories in task space, the Euclidean distance is informative enough for the trajectory comparison~\cite{stark_humanoids2017}. Thus, the Euclidean distance between the desired object trajectory $\textbf{T}^*$ and the known tasks $\textbf{T}_i$ is calculated 
$D_{\textbf{T}_i} = \norm{\textbf{T}^*-\textbf{T}_i}_2$.

We use the calculated distances and $k=1$, $k=2$, and $k=9$ as hyperparameter of the k-nearest neighbor algorithm to select the source knowledge. We combine these three modes with the partial transfer and with the full transfer. Together with the baseline, this yields seven different evaluation settings.\\

\subsubsection{Reward}
As due to the episode-based formulation of REPS, each sampled trajectory is executed at once, the reward is calculated only afterward for the whole sampled trajectory. The accumulated reward function for a trajectory $R(\boldsymbol{\tau}_i)$ consists of a component $r^\textbf{T}_i$ which evaluates how close the current object trajectory $\textbf{o}_i$ is to the target task descriptor $\textbf{T}^*$ to evaluate the quality of the current movement (and the resulting object trajectory). Again, the Euclidean distance is used to determine the similarity 
$r^\textbf{T}_i = \norm{\textbf{T}^*-\textbf{o}_i}_2$.  
To punish the robot when throwing the object instead of pushing it, we extend the reward function by a second component $r^p_i$ to reward pushing movements, which is formulated as the Euclidean distance between the end-effector trajectory $\textbf{e}_i$ and the object trajectory $\textbf{o}_i$ over the whole execution time, thus  
\begin{align}
	R(\boldsymbol{\tau}_i) & = a r^\textbf{T}_i + b r^p_i\nonumber                                                                                                                       \\
	  & = a\sqrt{(\textbf{T}^*-\textbf{o}_i)^{\intercal}(\textbf{T}^*-\textbf{o}_i)} + b \sqrt{(\textbf{e}_i-\textbf{o}_i)^\intercal(\textbf{e}_i-\textbf{o}_i)}. \nonumber
\end{align}
The factors $a$ and $b$ are hand-tuned such that the mean value of $r^\textbf{T}_i$ has $1.5$ times the value of the mean value of~$r^p_i$. While in theory, zero would be the optimal possible reward, due to taking the Euclidean distance over $1250$ time steps and also, the gripper not allowing for zero distance between the end-effector and the object, the robot is practically not able to actually gain a reward of zero.

\begin{figure*}
	\begin{minipage}[t]{\textwidth}
		\includegraphics{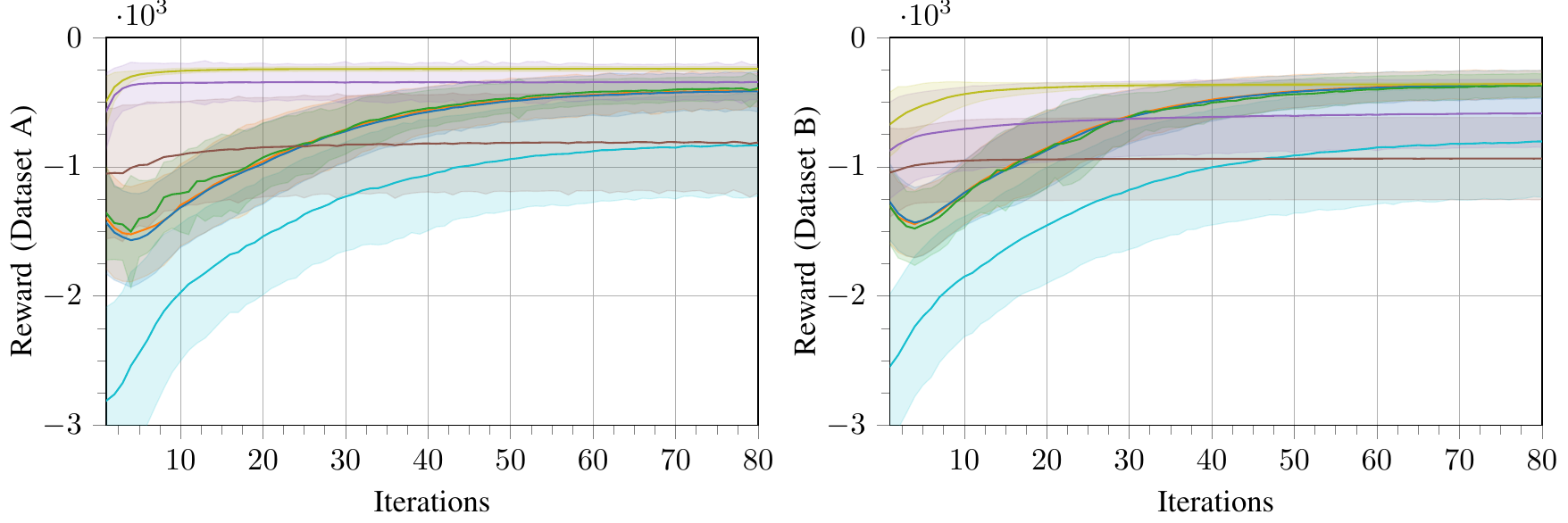}
	\end{minipage}
	\begin{minipage}[t]{\textwidth}
		\centering
		\vspace{-4mm}
		\includegraphics{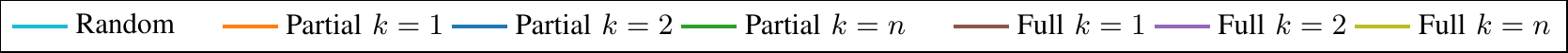}
	\end{minipage}
	\caption{Mean reward for each iteration, averaged over the learning of all ten different skills. The source knowledge is selected from all possible skill combinations, thus $N\in\{1,2,9\}$. The label $k=x$ denotes from how many source skills the initialization was compiled. The left plot shows the results for learning on dataset A, the right plot shows results for learning on dataset B. The plots include the standard deviation to indicate the variety of the results.}
	\label{allEvaluations}
\end{figure*}

\subsection{Evaluations}

In~\cite{torrey2010transfer}, three possible quantitative benefits are listed which can be gained from transfer learning: \textit{higher start}, \textit{higher asymptote}, and \textit{higher slope}. The first benefit is defined as the performance level gained by using only the transferred knowledge without any learning update. The second benefit is defined as a higher final performance level. The third one is defined as the time it takes to fully learn the target~task.

We evaluate all transfer possibilities between the skills, thus all possible source knowledge combinations are formed and used as initialization for learning each task which is not part of the source knowledge. Hence, we have settings in which the robot knows all skills and chooses the most appropriate one via k-NN, and we have settings in which there exists only a minimal library of exactly $k$ skill(s), i.e., the number of source skills the robot has to choose. The latter setting forces the robot to take suboptimal initializations as it has no actual choice when selecting the source knowledge. This is not the setting we have described within this paper, but settings which may occur during actual lifelong learning, as in such a scenario, not always the optimal set of source skills has been learned yet. 

We run REPS for $200$ iterations for all setups, but as all settings converge latest around iteration $60$, we show only $80$ iterations. In each iteration, $60$ parameter samples are drawn from the search distribution and each resulting trajectory is evaluated via the accumulated reward function $R(\boldsymbol{\tau})$. Each trajectory has a length of 1250 time steps and each time step takes $0.004$ seconds, resulting in $5$ seconds per trajectory execution.

Each of the transfer possibilities is evaluated five times. This results in overall $5\times 460 = 2300$ evaluations. For each evaluation of one of the settings, we also evaluate the baseline. The results for a large initial library allowing us to choose optimal $k$ source skills are provided in Figure~\ref{allMaxLibrary}. The results of enforcing suboptimal source knowledge are shown in Figure~\ref{allMinLibrary}. The average overall obtained results are shown~in~Figure~\ref{allEvaluations}. \\

\subsubsection{Evaluation On a Large Library}
Figure~\ref{allMaxLibrary} shows the average reward of the different transfer modalities for the learning of all skills from a large skill library. We can see that both introduced transfer approaches reach a higher asymptote than the baseline on almost all settings (except the full transfer with $k=1$ on dataset B). All full transfers reach their maximum fastest, within latest $25$ iterations, thus have the highest slope according to~\cite{torrey2010transfer} and learn with the least amount of required samples. The partial transfer outperforms the full transfer approach in the setting of $k=1$. The convergence takes approximately $60$ iterations, which is about the same number of iterations as the baseline. Regarding the highest start criterion, both transfer approaches yield the same initial reward as the reward is calculated based on the mean~only.  \\

\begin{figure*}
	\includegraphics[width=\textwidth]{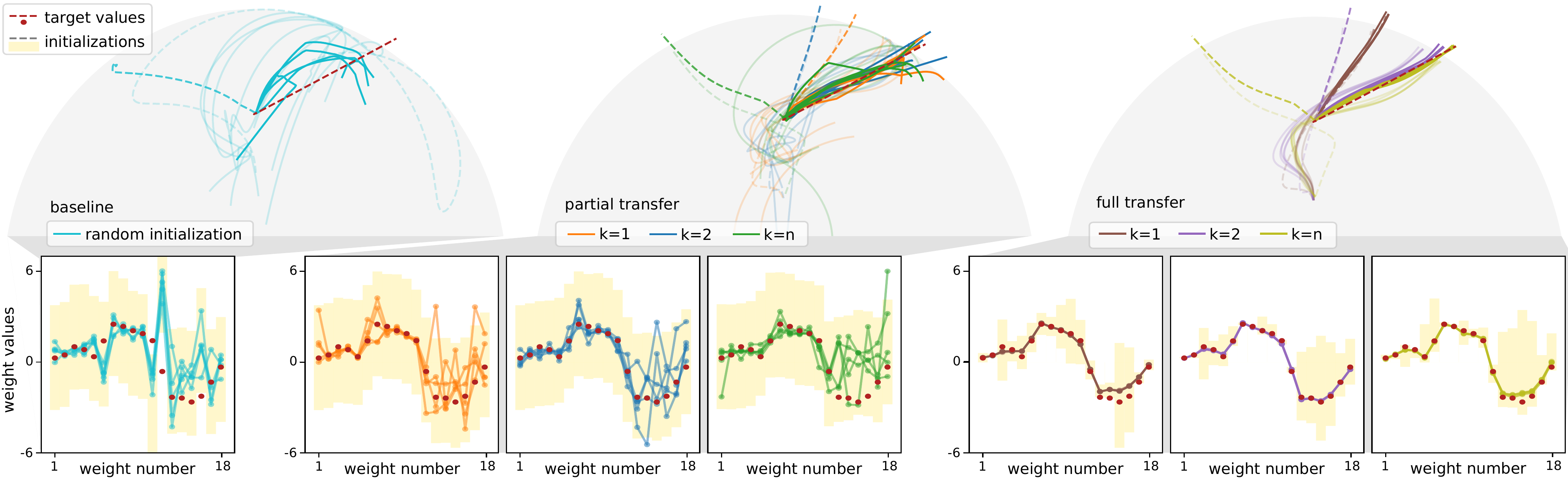}
	\caption{To grasp the qualitative results of the transfer learning approaches, we show exemplary resulting behavior of five learning trials for each of the different transfer strategies on training set B. The upper row shows the workspace of the robot, the light trajectories depict the end-effector movements, the dark lines show the object trajectories and the dashed lines show the initializations. In the lower row, the yellow bars depict the initial search space (standard deviation) of each parameter of the movement trajectory. The red dots show one possible solution to the task which has been used to generate the dataset. For the random initialization on the very left, the resulting skills were rather random movements resulting in mediocre obstacle trajectories. For the partial transfer (middle), the resulting object trajectories improved but the end-effector trajectories still deviate a lot from the initial movement and thus have unpredictable behavior. For full transfer (right), the resulting trajectories were similar to the source knowledge, though, in the case of $k=1$, transfer usually did not take place at all.}
	%\vspace{-10pt}
	\label{figQualitativeEvaluation}
\end{figure*}

\subsubsection{Evaluation on a Small Library}
For a small library not allowing optimal source knowledge, the transfer results are shown in Figure~\ref{allMinLibrary}. For almost all settings, the partial transfer yields the highest asymptote and it always clearly outperforms the baseline. The full transfer approach still has the shortest slope and thus the least amount of required samples, but it only excels on dataset A and $k=2$. For the other cases, even the baseline reaches a higher reward level. This shows the sensitivity of the full transfer approach regarding the quality of the selected source knowledge. Thus, the full transfer benefits from a large library to select source knowledge from.

 Figure~\ref{allEvaluations} shows all of the evaluated transfers in one plot and further emphasizes how much the full transfer depends on good source knowledge, as in overall average, full transfer is only able to reliably reach the same asymptote as the partial transfer when using $k=9$ source skills. In contrast, the results of the partial transfer are invariant regarding the amount of library skills the source knowledge is selected~from.\\

\subsubsection{Qualitative Evaluation}
Though the reward values shown in Figure~\ref{allMaxLibrary} indicate that the robot learns to push the object closely along the target trajectory in most of the cases, there are qualitative differences between the solutions which we depict in Figure~\ref{figQualitativeEvaluation}. 
We visualize these qualitative differences by comparing the end-effector trajectories. Especially for dataset B, where the robot arm has to learn how to approach the object, the different shapes of the solutions are obvious: when starting only from a source knowledge mean or from a random initialization, the results are chaotic end-effector movements and resemble often a hitting or throwing movement instead of a gentle pushing. In contrast, for an initialization with a full transfer, the shape of the trajectory is preserved. Thus, the learned movements are similar to the already known ones and by this are more predictable and safer (as they do not enter dangerous areas) than the other initialization possibilities. 

Thus, Figure~\ref{allMaxLibrary} and Figure~\ref{figQualitativeEvaluation} together show that there is a trade-off when transferring knowledge. One can either learn a specific, predictable solution with small variance regarding the learned trajectory and high similarity to its initialization, yielded by using full transfer. Alternatively, one can learn a solution with a less similar movement trajectory but guaranteed to achieve a high reward, thus the objectively better solution to the provided task. The latter is yielded by using partial transfer.\\

\subsubsection{Discussion}
Within the presented evaluations are two observations which are contrary to (at least our) intuition and thus may be from interest. First, when using full transfer, taking only the most similar skill as initialization ($k=1$) does not always allow a good transfer. It happens that the parameter search starts in a local optimum and has too little exploration space which it would need to leave the initial optimum. A larger scaling factor $s$ for the covariance matrix did not solve this problem, which means that the shape of the search space is important. This is especially the case for the training dataset B, where the robot starts with a distance to the object and thus is unlikely to actually touch the object when executing random movements. In some cases, adding terms such as novelty, or aversion, towards the initial solution to the cost function help overcoming this problem, but they also tend to move the solution away from the pushing movement we are aiming at towards movements hitting or throwing the cylinder.
This observed transfer behavior (but not necessarily the internal process) for $k=1$ in the full transfer show an interesting similarity between our framework and actual human learning: for humans, a negative transfer appears usually in tasks which have high perceptual similarity~\cite{Osgood1949-OSGTSP} as this circumstance allows for confusion~\cite{Annett1985}.
Furthermore, this also relates to humans having difficulties to unlearn already learned mistakes from earlier stages of training~\cite{Annett1985}, though our framework is nowhere near to taking anything into account like muscle memory or~habits.

Second, when the source knowledge is selected from a large skill library, it happens that for $k=2$, the mean of the two closest skills gives an already almost perfect initialization, as an interpolation between the provided tasks works sufficiently for the 3DOF robot. Interestingly, those initializations do not always gain the best final reward. Often, a seemingly worse (because more general and thus further away) initialization of $k=n$ yields a better final reward, which can be seen in Figure~\ref{allMaxLibrary}. This is most likely due to a search distribution being gained from more as well as wider source knowledge, providing fewer restrictions.

\section{Conclusion}

In this paper, we showed that a deliberate choice of the parameter search space for a learning algorithm working on a search distribution enables safer, more predictable and faster learning of a new motor skill, which all are favorable properties for skill learning in robotics. We showed that the selection of such an appropriate initial search distribution can be done by transfer learning as demonstrated within this paper. We proposed two different approaches for transferring of already gained knowledge to the new task. Depending on the overall aim, there are different criteria according to which we recommend respective transfer approaches: in case of only the solution of the task being important, independently of how the solution may be executed, then simply transferring the mean of the source search distribution enables the robot to find solutions which may be more accurate but also further away from the skills it knows already. For the case of wanting safe exploration on the robot or movements which are similar to the ones the robot knows already, the full transfer approach is the initialization to go with. It is also our recommendation for setups where evaluations are hard to obtain, as the full transfer needs the least amount of learning samples. 

The ideal setting for full transfer is to have as many as possible similar skills of which the source knowledge can be combined. In this case, the full transfer yields an initialization balanced between providing useful knowledge to the robot and yet being not too restricting on the exploration space.
Further investigations on how to provide optimal source knowledge for the full transfer could enable faster learning on the robot itself, as a well initialized full transfer saves about $60\%$ of samples. Thus, it might be fruitful to investigate into finding optimal orders of skill learning to always provide the best possible source knowledge to the~robot. 

Our approach is one step towards enabling future intelligent, adaptive and thus autonomous robots which are capable of lifelong learning. Such robots could be delivered with a small set of basic skills that allows them to rapidly build their skill library of required skills to deal with their environment. Such a basic skill set also enables simple teaching by non-expert humans, who have to demonstrate only a few desired trajectories and then can let the robot learn by itself without having to supervise it closely, or, can demonstrate skills which are easy to demonstrate but not actually solve the desired task and then let the robot improve the demonstration.

\addtolength{\textheight}{-9.20cm}

\bibliographystyle{IEEEtran}
\bibliography{references}

\begin{thebibliography}{10}

\bibitem{adolph2017development}
K.~E. Adolph and J.~M. Franchak, ``The development of motor behavior,'' {\em
  Wiley Interdisciplinary Reviews: Cognitive Science}, 2017.

\bibitem{NIPS2013_5177}
A.~Paraschos, C.~Daniel, J.~R. Peters, and G.~Neumann, ``Probabilistic movement
  primitives,'' in {\em Advances in Neural Information Processing Systems 26},
  2013.

\bibitem{Taylor:2009:TLR:1577069.1755839}
M.~E. Taylor and P.~Stone, ``Transfer learning for reinforcement learning
  domains: A survey,'' {\em The Journal of Machine Learning Research}, 2009.

\bibitem{Weiss2016}
K.~Weiss, T.~M. Khoshgoftaar, and D.~Wang, ``A survey of transfer learning,''
  {\em Journal of Big Data}, 2016.

\bibitem{5288526}
S.~J. {Pan} and Q.~{Yang}, ``A survey on transfer learning,'' {\em IEEE
  Transactions on Knowledge and Data Engineering}, 2010.

\bibitem{HASSABIS2017245}
D.~Hassabis, D.~Kumaran, C.~Summerfield, and M.~Botvinick,
  ``Neuroscience-inspired artificial intelligence,'' {\em Neuron}, 2017.

\bibitem{DBLP:journals/corr/abs-1808-01974}
C.~Tan, F.~Sun, T.~Kong, W.~Zhang, C.~Yang, and C.~Liu, ``A survey on deep
  transfer learning,'' in {\em International conference on artificial neural
  networks}, Springer, 2018.

\bibitem{DBLP:journals/corr/LakeUTG16}
B.~M. Lake, T.~D. Ullman, J.~B. Tenenbaum, and S.~J. Gershman, ``Building
  machines that learn and think like people,'' {\em Behavioral and brain
  sciences}, 2017.

\bibitem{youtube}
B.-H. Kim, ``Nips 2016 tutorial: Nuts and bolts of building ai applications
  using deep learning by andrew ng.''
  https://www.youtube.com/watch?v=wjqaz6m42wU.
\newblock 2019-02-13.

\bibitem{DBLP:journals/corr/HelwaS17}
M.~K. Helwa and A.~P. Schoellig, ``Multi-robot transfer learning: A dynamical
  system perspective,'' in {\em 2017 IEEE/RSJ International Conference on
  Intelligent Robots and Systems (IROS)}, IEEE, 2017.

\bibitem{Makondo2018AcceleratingML}
N.~Makondo, B.~Rosman, and O.~Hasegawa, ``Accelerating model learning with
  inter-robot knowledge transfer,'' {\em 2018 IEEE International Conference on
  Robotics and Automation (ICRA)}, 2018.

\bibitem{DBLP:journals/corr/abs-1809-04720}
J.~van Baar, A.~Sullivan, R.~Cordorel, D.~Jha, D.~Romeres, and D.~Nikovski,
  ``Sim-to-real transfer learning using robustified controllers in robotic
  tasks involving complex dynamics,'' in {\em 2019 International Conference on
  Robotics and Automation (ICRA)}, IEEE, 2019.

\bibitem{Schwab2018ZeroST}
D.~Schwab, Y.~Zhu, and M.~M. Veloso, ``Zero shot transfer learning for robot
  soccer,'' in {\em International Conference on Autonomous Agents and
  Multiagent Systems (AAMAS)}, 2018.

\bibitem{AAMASWS10-barrett}
S.~Barrett, M.~E. Taylor, and P.~Stone, ``Transfer learning for reinforcement
  learning on a physical robot,'' in {\em Ninth International Conference on
  Autonomous Agents and Multiagent Systems - Adaptive Learning Agents Workshop
  (AAMAS - ALA)}, 2010.

\bibitem{Fernandez:2006:PPR:1160633.1160762}
F.~Fern\'{a}ndez and M.~Veloso, ``Probabilistic policy reuse in a reinforcement
  learning agent,'' in {\em Proceedings of the Fifth International Joint
  Conference on Autonomous Agents and Multiagent Systems (AAMAS)}, 2006.

\bibitem{torrey2010transfer}
L.~Torrey and J.~Shavlik, ``Transfer learning,'' in {\em Handbook of Research
  on Machine Learning Applications and Trends: Algorithms, Methods, and
  Techniques}, 2010.

\bibitem{Taylor06transferlearning}
M.~E. Taylor, S.~Whiteson, and P.~Stone, ``Transfer learning for policy search
  methods,'' in {\em In ICML Workshop on Structural Knowledge Transfer for
  Machine Learning}, 2006.

\bibitem{7781959}
A.~{Abdolmaleki}, D.~{Simões}, N.~{Lau}, L.~P. {Reis}, and G.~{Neumann},
  ``Contextual relative entropy policy search with covariance matrix
  adaptation,'' in {\em 2016 International Conference on Autonomous Robot
  Systems and Competitions (ICARSC)}, 2016.

\bibitem{promps_auro}
A.~Paraschos, C.~Daniel, J.~Peters, and G.~Neumann, ``Using probabilistic
  movement primitives in robotics,'' {\em Autonomous Robots}, 2018.

\bibitem{Rueckert_ICRA14LMProMPsFinal}
E.~Rueckert, J.~Mundo, A.~Paraschos, J.~Peters, and G.~Neumann, ``Extracting
  low-dimensional control variables for movement primitives,'' in {\em
  Proceedings of the International Conference on Robotics and Automation
  (ICRA)}, 2015.

\bibitem{srep28455}
E.~Rueckert, J.~Camernik, J.~Peters, and J.~Babic, ``Probabilistic movement
  models show that postural control precedes and predicts volitional motor
  control,'' {\em Nature PG: Scientific Reports}, 2016.

\bibitem{6439}
J.~Peters, K.~M{\"u}lling, and Y.~Altun, ``Relative entropy policy search,''
  {\em Proceedings of the Twenty-Fourth National Conference on Artificial
  Intelligence}, 2010.

\bibitem{Deisenroth:2013:SPS:2688186.2688187}
M.~P. Deisenroth, G.~Neumann, and J.~Peters, ``A survey on policy search for
  robotics,'' {\em Foundations and Trends in Robotics}, 2013.

\bibitem{wierstra2014natural}
D.~Wierstra, T.~Schaul, T.~Glasmachers, Y.~Sun, J.~Peters, and J.~Schmidhuber,
  ``Natural evolution strategies.,'' {\em Journal of Machine Learning
  Research}, 2014.

\bibitem{hansen1996adapting}
N.~Hansen and A.~Ostermeier, ``Adapting arbitrary normal mutation distributions
  in evolution strategies: The covariance matrix adaptation,'' in {\em
  Proceedings of IEEE international conference on evolutionary computation},
  1996.

\bibitem{mania2018}
H.~Mania, A.~Guy, and B.~Recht, ``Simple random search of static linear
  policies is competitive for reinforcement learning,'' in {\em Advances in
  Neural Information Processing Systems 31}, 2018.

\bibitem{doi:10.1080/00031305.1992.10475879}
N.~S. Altman, ``An introduction to kernel and nearest-neighbor nonparametric
  regression,'' {\em The American Statistician}, 1992.

\bibitem{Ijspeert:2013:DMP:2432779.2432781}
A.~J. Ijspeert, J.~Nakanishi, H.~Hoffmann, P.~Pastor, and S.~Schaal,
  ``Dynamical movement primitives: Learning attractor models for motor
  behaviors,'' {\em Neural Computation}, 2013.

\bibitem{coumans2018}
E.~Coumans and Y.~Bai, ``Pybullet, a python module for physics simulation for
  games, robotics and machine learning.'' \url{http://pybullet.org},
  2016--2018.

\bibitem{stark_humanoids2017}
S.~Stark, J.~Peters, and E.~Rueckert, ``A comparison of distance measures for
  learning nonparametric motor skill libraries,'' in {\em Proceedings of the
  International Conference on Humanoid Robots (HUMANOIDS)}, 2017.

\bibitem{Osgood1949-OSGTSP}
C.~E. Osgood, ``The similarity paradox in human learning: A resolution,'' {\em
  Psychological Review}, 1949.

\bibitem{Annett1985}
J.~Annett and J.~Sparrow, ``Transfer of training: A review of research and
  practical implications,'' {\em Innovations in Education and Training
  International}, 1985.

\end{thebibliography}

\end{document}